\documentclass[letterpaper, 10 pt, conference]{ieeeconf}  % Comment this line out if you need a4paper
\usepackage{graphicx}

\usepackage{amsmath}
\usepackage{amsfonts}
\usepackage{xcolor}
\usepackage{tabularx}
\usepackage{makecell}
\usepackage{booktabs} 
\IEEEoverridecommandlockouts                              % This command is only needed if 
\title{\LARGE \bf
Explaining Autonomous Driving Actions with Visual Question Answering
}

\author {Shahin Atakishiyev$^{1}$, Mohammad Salameh$^{2}$, Housam Babiker$^{1}$, Randy Goebel$^{1}$ \thanks{ This work was supported by 
 the Alberta Machine Intelligence Institute (Amii),  Computing Science Department of the University of Alberta, and the Natural Sciences and Engineering Research Council of Canada (NSERC).}% <-this % stops a space
\thanks{Shahin Atakishiyev, Housam Babiker, and Randy Goebel are with the Department of Computing Science, University of Alberta, Edmonton, Alberta, Canada.
}% <-this % stops a space
\thanks{Mohammad Salameh is with Huawei Technologies Canada Co., Ltd., Edmonton, Alberta, Canada.}% <-this % stops a space
\thanks{Correspondence: \tt\small shahin.atakishiyev@ualberta.ca}% <-this % stops a space
}

\begin{document}

\maketitle
\thispagestyle{empty}
\pagestyle{empty}

%%%%%%%%%%%%%%%%%%%%%%%%%%%%%%%%%%%%%%%%%%%%%%%%%%%%%%%%%%%%%%%%%%%%%%%%%%%%%%%%
\begin{abstract}

The end-to-end learning ability of self-driving vehicles has achieved significant milestones over the last decade owing to rapid advances in deep learning and computer vision algorithms. However, as autonomous driving technology is a safety-critical application of artificial intelligence (AI), road accidents and established regulatory principles necessitate the need for the explainability of intelligent action choices for self-driving vehicles. To facilitate interpretability of decision-making in autonomous driving, we present a Visual Question Answering (VQA) framework, which explains driving actions with question-answering-based causal reasoning. To do so, we first collect driving videos in a simulation environment using reinforcement learning (RL) and extract consecutive frames from this log data uniformly for five selected action categories. Further, we manually annotate the extracted frames using question-answer pairs as justifications for the actions chosen in each scenario. Finally, we evaluate the correctness of the VQA-predicted answers for actions on unseen driving scenes. The empirical results suggest that the VQA mechanism can provide support to interpret real-time decisions of autonomous vehicles and help enhance overall driving safety.

\end{abstract}

%%%%%%%%%%%%%%%%%%%%%%%%%%%%%%%%%%%%%%%%%%%%%%%%%%%%%%%%%%%%%%%%%%%%%%%%%%%%%%%%
\section{INTRODUCTION}
Urban autonomous driving is one of the most challenging tasks for self-driving vehicles, especially considering the potential interaction with other cars, road-crossing pedestrians, bystanders, traffic lights, and other conditions of dynamically changing environments. As highly automated vehicles increasingly rely on mapping sensory data to control the commands of a vehicle, applicable end-to-end learning techniques should be acceptably safe and computationally transparent. In particular, the remarkable success of deep learning and computer vision algorithms has expedited progress in safe autonomous driving on real roads and urban areas. For example, in February 2023, Waymo reported that their autonomous vehicle drove more than one million rider-only miles across several US cities with no reported injuries or events involving vulnerable road participants \cite{waymo_one_million}. The report also describes that Waymo's vehicle was involved in two accidents, where one of the accidents was caused by the driver of another car being distracted by their phone while approaching a red traffic light, according to Waymo's claim. Moreover, other recently reported traffic accidents
\begin{figure}[t!]
\vspace{0.30cm}
\centering
\includegraphics[width = 1\columnwidth]{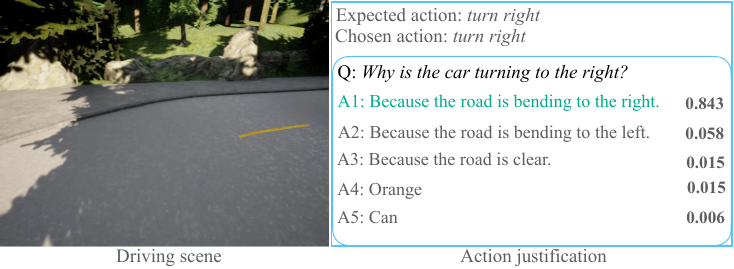}
\caption{An example of the most probable answers with softmax probability scores predicted by our VQA framework on the action of an ego vehicle.}
\label{fig:action_justification}
\end{figure}
with self-driving cars \cite{board2020collision} and resulting fatalities call for a scrutinized regulation of vehicle autonomy within a legal framework. Such road mishaps trigger safety, transparency, and other legal culpability issues. Inherently, a self-driving vehicle also needs to justify its temporal decisions with some form of explanation.
 As self-driving decisions directly impact passengers on board and other road users, consumers and transportation jurisdictions intrinsically expect transparency and rely on the correctness of such decisions. As a concrete example, the European Union (EU) adopted the General Data Protection Regulation (GDPR) \cite{voigt2017eu} that proposed a recital of the \textit{right of explanation}, which entitles consumers to receive an explanation on decisions of autonomous systems. Article 22 of GDPR also describes general principles regarding stakeholders' rights and responsibilities to use automated decision-making systems \cite{gdpr_article_22}. Thus, the need for the explainability of autonomous driving decisions has legal, socio-technical, psychological, and philosophical perspectives, in general \cite{omeiza2021explanations, atakishiyev2021explainable}.\\
The delivery of explanations is another important topic in autonomous driving. As both consumers and engaged technical people have different backgrounds and knowledge about how self-driving cars work, it is necessary that explanations are provided in accordance with an explanation receiver's (i.e., an explainee) relevant identity, as described in the recent surveys of \cite{omeiza2021explanations, atakishiyev2021explainable, zablocki2022explainability}. In this context, self-driving explanations must be correct, sufficiently informative, and intelligible for explainees. \\
In this study, we propose a VQA-based explanation approach\footnote{The source code, data, and related resources are available at https://github.com/Shahin-01/VQA-AD.} to justify RL-based self-driving decisions in a simulation environment. At its core, VQA is a task in the intersection of natural language processing and computer

\begin{figure*}[htp!]
 \centering
    \includegraphics[width=17.5 cm]{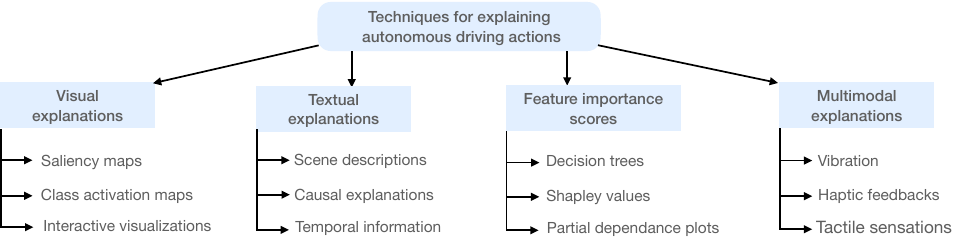}
    \caption{The most common approaches to explaining autonomous driving actions.}
    \label{fig:exp}
\end{figure*}\hspace{-0.50cm}
vision, which produces an answer to a text-based question about an image \cite{kafle2017visual}. Such an objective makes this reasoning technique intuitively applicable to autonomous driving. When humans drive or are a passenger on board, they inherently analyze real-time and upcoming traffic scenes and think about relevant causal, temporal, and descriptive questions, such as ``Why is the car turning left?'', ``What action will the car in the left lane perform at the T-junction?'' and ``What is the speed of the vehicle in front?'' as examples. Getting answers to such questions by any means helps us have a reliable and safe trip. In this regard, we leverage the VQA mechanism to pose a question on an autonomous car’s chosen action within the driving scene and justify the question with a causal answer reflecting the car’s decision-making in that scenario. \\ 
 Motivated by this point, we build our framework as follows. We train an RL agent (i.e., an ego car) to operate in an autonomous driving environment and record its decisions (\textit{actions}) in correspondence to the video frames (\textit{states}). We then utilize a VQA system to justify actions of the ego car: the VQA framework inputs an \textit{image frame}  with a \textit{question} reflecting the action of the car in the scene and tries to predict the relevant answer for such an action (e.g., Figure \ref{fig:action_justification}).   \\ 
Overall, the main contributions of our paper can be summarized as follows:
\begin{itemize}
    \item We present the first empirical study on explaining autonomous driving actions with a VQA approach.
    \item We release a dataset of image-question-answer triplets justifying an autonomous car’s actions in the scene.
    \item We show that connecting vision and natural language could rationalize an RL agent's decision-making in an intelligible way.
    \item We propose further directions to develop  more rigorous VQA frameworks for explanatory self-driving actions.
\end{itemize}
 The rest of the paper is organized as follows. In Section II, we provide state-of-the-art explainability approaches for autonomous driving. We then present details of the data generated by our RL agent and visual feature extraction in Section III. Finally, we report empirical results and the discussion of these results in Section IV and sum up the article with concluding remarks and future directions.

\section{Related Work}

Since Bojarski et al.'s \cite{bojarski2016end} CNN-based end-to-end learning approach, the autonomous driving community has shown increasing interest in interpreting self-driving decisions. In general, primarily explored explanation provision techniques for autonomous driving are \textit{visual, textual explanations},  \textit{feature importance scores}, and  \textit{hybrid} or \textit{multimodal explanations} comprising two or more of these methods (see Figure ~\ref{fig:exp} for the relevant classification). \\
Visual explanations in the context of autonomous driving identify which parts of the perceived image (i.e., driving scene) have more influence on the vehicle’s decision, as justifications for the performed action \cite{zablocki2022explainability}. For instance, a visual explanation can show an image of a red traffic light captured by the vehicle’s video camera as a \textit{saliency map} (i.e., a heatmap) pointing out that the perception algorithm classified it as a primary reason for stopping. In this context, Kim and Canny proposed a causal attention-based visualization technique to show which groups of pixel values (i.e., blobs) have a true causal impact on the model's prediction \cite{kim2017interpretable}. After analyzing attention maps in a post-hoc manner, they remove more than half of the blobs and analyze the model's output. The empirical results show that the network produces more convincing and correct predictions in driving decisions, just like real drivers do in a realistic environment. Furthermore, as an augmented version of their initial work \cite{bojarski2016end}, Bojarski et al. developed \textit{VisualBackProp}, a saliency map-based visual explanation framework highlighting which sets of input pixels have more influence on a vehicle's decisions \cite{bojarski2018visualbackprop}. They show that the VisualBackProp method correctly identifies the most important traffic elements, such as lane markings and other  cars in the scene, as a basis for decision-making. In addition, VisualBackProp has been proven to be an effective interpretable approach to detecting the failure cases \cite{mohseni2019predicting} in the original vision-based end-to-end learning method of {\cite{bojarski2016end}.\\     
Another popular vision-based rationalization technique uses the idea of \textit{counterfactual visual explanations}. These explanations aim to identify whether changing some parts of the original image leads to a different prediction than the original prediction made on the original input. Bansal et al. \cite{bansal2019chauffeurnet} modified hand-crafted inputs by removing some objects in the image to see whether their introduced \textit{ChauffeurNet} makes different predictions with the altered image. A similar strategy is followed by Li et al. \cite{li2020make}, where the goal is to find the \textit{risk objects} for driving. They show that manipulated

\begin{figure}[t!]
\includegraphics[width = \columnwidth]{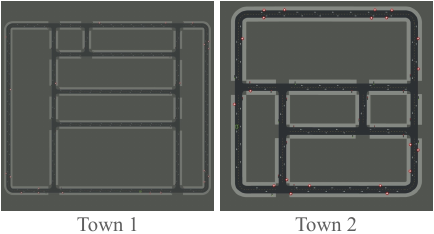}
\caption{An aerial view of Town 1 and 2 on the CARLA simulator \cite{dosovitskiy2017carla}.}
\label{fig:towns}
\end{figure}
\hspace{-0.45cm}
removal of a pedestrian in an intersection changes the driving command from ``Stop'' to ``Go''; thus, the pedestrian is considered a ``risk object'', which causes  the driving decision to change to the contrastive class. Finally, as a more recent counterfactual analysis, Jacob et al. \cite{jacob2022steex} investigated a style modification of image regions on the driving model's predictions. The experimental study shows that their presented framework, \textit{STEEX}, generates counterfactual explanations in case of manual interventions to the driving scene. Therefore, visual explanations can enable people to ensure that the intelligent driving system  accurately senses the operational environment. \\  
Textual descriptions are another way of conveying rationales  to the end-users for driving decisions. This approach generates natural language text that explains driving actions with descriptive, temporal, and causal information. The first successful textual explanation work is Kim et al.'s study \cite{kim2018textual}, where the authors leverage an attention-based video-to-text approach to generate textual explanations for autonomous vehicles. They further extend this work by incorporating human advice \cite{kim2019grounding} and observation-to-action rules \cite{kim2020advisable} to the underlying model and provide text-based explanations on performed actions. In another study, Xu et al. introduce \textit{BDD-OIA }\cite{xu2020explainable}, an extension of the BDD100K dataset. Based on the action-inducing objects, they provide 21 explanations for a set of 4 actions (move forward, stop/slow down, turn left, turn right). Lastly, in this context, Ben Younes et al. \cite{ben2022driving} proposed \textit{BEEF}, an architecture that explains the behavior of trajectory prediction with textual justifications based on features fused from multi-levels, such as late features comprising the system-wise decisions and spatio-temporal features consisting of perceptual driving information. \\
Feature importance scores, as well-known quantitative evaluation metrics, have also recently been investigated in various autonomous driving tasks. The applications of these methods to autonomous driving include decision trees \cite{brewitt2021grit}, Shapley values \cite{ almalioglu2022deep}, and partial dependence plots \cite{lee2019attitudes}. The primary goal of these methods in self-driving is to understand the weights and contributions of scene features used in predictive modeling across the explored self-driving tasks. \\
Finally, except for visual, textual, and quantitative explanations, recent studies have attempted to use multi-modal explanatory  techniques to convey information on the chosen course of actions of self-driving vehicles. For example, in

\begin{figure}[t!]
\centering
\includegraphics[width = 0.625\columnwidth]{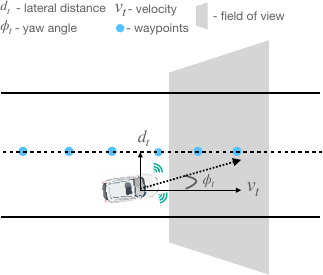}
\caption{State space representation of the ego car in its environment. An ideal state is that the car follows the direction of the lane within the lane.}
\label{fig:state_space}
\end{figure} \hspace{-0.52cm}
their two studies \cite{schneider2021increasing, schneider2021explain}, Schneider et al. propose a combination of visual, textual, audio, light, and vibration feedback to provide retrospective and live explanations on action decisions of autonomous driving. These studies show that while visualization and light-based driving information improve the user experience (UX), multi-modal explanations can enhance perceived control and understanding of a vehicle's decision-making by connecting UX, autonomous driving, and explainable AI. Moreover, real-time driving information delivered via vibration, tactile sensation, or haptic feedback with a relevant degree of an alert may have a crucial role in the smooth and timely transfer of control between a self-driving vehicle and a backup driver. \\
With the inherent ability to reason about visual information, such as images, videos, and related multimedia data,  VQA has recently been explored in several safety-critical and security-concerning domains. These works include applications to the healthcare and medical field \cite{gurari2018vizwiz, zhan2020medical} and visual surveillance \cite{li2019isee}. Interestingly, the topic has not been investigated deeply in autonomous driving. As far as we know, there are only two instances that utilize the VQA mechanism in the transportation domain. The first is the CLEVRER dataset, which describes the collision events with video representation and reasoning \cite{Yi2020CLEVRER}. The other contribution is the SUTD-TrafficQA benchmark, which basically predicts traffic situations with question-answer pairs ranging from basic understanding (i.e., \textit{What is the type of the road?}) to reverse reasoning (i.e., \textit{What might have happened moments ago?}) \cite{xu2021sutd}. In our study, we focus on action-based explanations and show that question-answering-based causal event reasoning has significant benefits for explaining real-time decisions of self-driving cars. We describe the details of the framework and the experimental results in the following sections. 

\section{Experimental Design and Methodology}
% Our proposed framework has three primary constituents.
%These are data collection, data annotation, and question-answering components. The details of these components and
%their roles are described in the following subsections for an
%easy understanding of the presented architecture.%
Our framework is designed in three primary steps. First, we use a deep RL agent to control an autonomous car in a simulation environment and collect a driving video from
\begin{table*}
  \centering
  \caption{Annotated question-answer pairs in our VQA framework}
  \label{tab:my_table}
  \begin{tabularx}{\linewidth}{cXc}
    \toprule
    \textbf{Action category} & \textbf{Question} & \textbf{Answer} \\
    \midrule
    Go straight & Why is the car going straight? & Because the road is clear.\\
    \addlinespace
    Turn left & Why is the car turning to the left? & Because the road is bending to the left. \\
    \addlinespace
    Turn left at T-junction & Why is the car turning left at T-junction? & Because there is no obstacle on the right side and turning left can be performed safely.  \\
    \addlinespace
    Turn right & Why is the car turning to the right? & Because the road is bending to the right. \\
    \addlinespace
    Turn right at T-junction & Why is the car turning right at T-junction? & Because there is no obstacle on the left side and turning right can be performed safely. \\
    \bottomrule
  \end{tabularx}
  \label{Tab:Tcr}
\end{table*}
\begin{figure}[t!]
\centering
\includegraphics[width = 1\columnwidth]{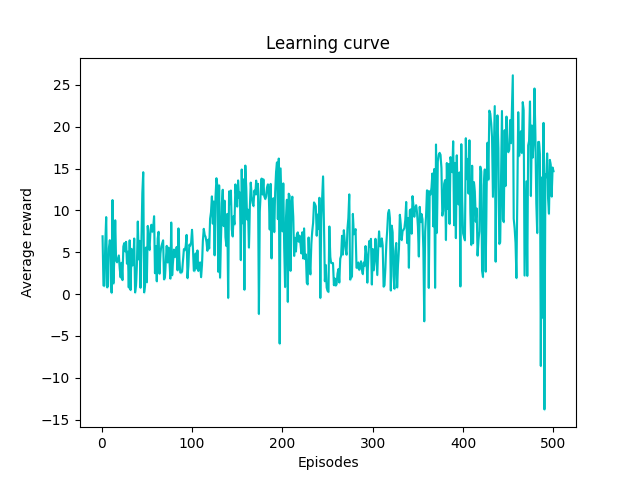}
\caption{Learning curve of DDPG in Town 1 with the specified parameters. Our VQA framework is further fine-tuned on driving data collected here.}
\label{fig:learning_curve}
\end{figure}\hspace{-0.1cm}
the simulator. We then convert this recorded video to image sequences uniformly. Finally, we select five specific action categories in the extracted driving frames and annotate them using question-answer pairs that justify the car’s action in the scene (Table \ref{Tab:Tcr}). The high-level description of the components and overall architecture is provided in Figure \ref{fig:VQA_AD}. Given such a setup, the objective of our architecture is to predict the correct answer to a posed question about an autonomous car’s performed action in an unseen driving scene.  The details of the data collection,  data annotation, and question-answering steps are described in the following subsections.

\subsubsection{Data Collection}To obtain driving data, we trained an RL agent (i.e., a self-driving car) on the CARLA simulator \cite{dosovitskiy2017carla}. We used the Deep Deterministic Policy Gradient (DDPG) algorithm  \cite{lillicrap2016} for the control of a self-driving car in a simulation environment. Control commands of automated driving have continuous actions including braking, acceleration, and steering angle which themselves can have a broad range of values as a representation. DDPG, as an augmented version of the Deep Q-learning algorithm, is particularly well-adapted for continuous action spaces and therefore is appropriate for driving control tasks. Furthermore, DDPG uses \textit{experience replay}, a memory storing the agent's past experiences ($s_t, a_t, r_t, s_{t+1}$), out of which the algorithm can sample randomly to train the agent. This ability to reuse samples makes DDPG a computationally efficient learning approach. Moreover, DDPG has an actor-critic architecture, in which the actor learns an observation-to-action mapping, and the critic learns to evaluate the quality of an agent's chosen actions. DDPG also uses \textit{target networks} - the target actor network $\mu'$, and target critic networks $Q'$. These networks are time-delayed copies of their original networks that help stabilize the training process. The parameters of target networks are updated as follows:

\begin{table}
  \centering
  \caption{The training parameters of DDPG on CARLA}
  \label{tab:my_table}
  \begin{tabularx}{\linewidth}{X X X X X X}
    \hline
    Actor learning rate & Critic learning rate & Target network hyper-parameter & Replay buffer size & Batch size & Discount factor\\
    \hline
    0.0001 & 0.001 & 0.001 & 100000 & 32 & 0.99 \\
    \hline
  \end{tabularx}
  \label{Tab:tr_params}
\end{table}
\hspace{-0.50cm}

\begin{equation}
            \theta^{Q'} \leftarrow \tau \theta^{Q} + (1 - \tau) \theta^{Q'}
          \end{equation}
          
\begin{equation}
            \theta^{\mu'} \leftarrow \tau \theta^{\mu} +
                (1 - \tau) \theta^{\mu'}
\end{equation}

where $\tau$ $\ll$ 1. For an effective action exploration, the term additive noise is usually added to the exploration policy and action is selected accordingly:

\begin{equation}
    a_t = \mu(s_t | \theta^{\mu}) + \mathcal{N}_t
\end{equation}
Such a learning technique enables the DDPG agent to learn a policy that maximizes its expected reward while also considering  the quality of the chosen actions. \\
\textit{RL Training Details:} We generated driving data by training the agent on Town 1 within CARLA. Town 1 (see Figure \ref{fig:towns}, a) is a map containing straight lines, left turns, right turns, T-junctions, traffic lights, speed signs, and various stationary objects around the curbs. We first use the A* motion planning algorithm \cite{hart1968formal} to generate a route with an initial and final point of a motion trajectory inside the simulated town, which shows consecutive waypoints linking these points. In our experiment, we set the number of waypoints to 15. By

\begin{figure*}[htp!]
 \centering
    \includegraphics[width=17.5 cm]{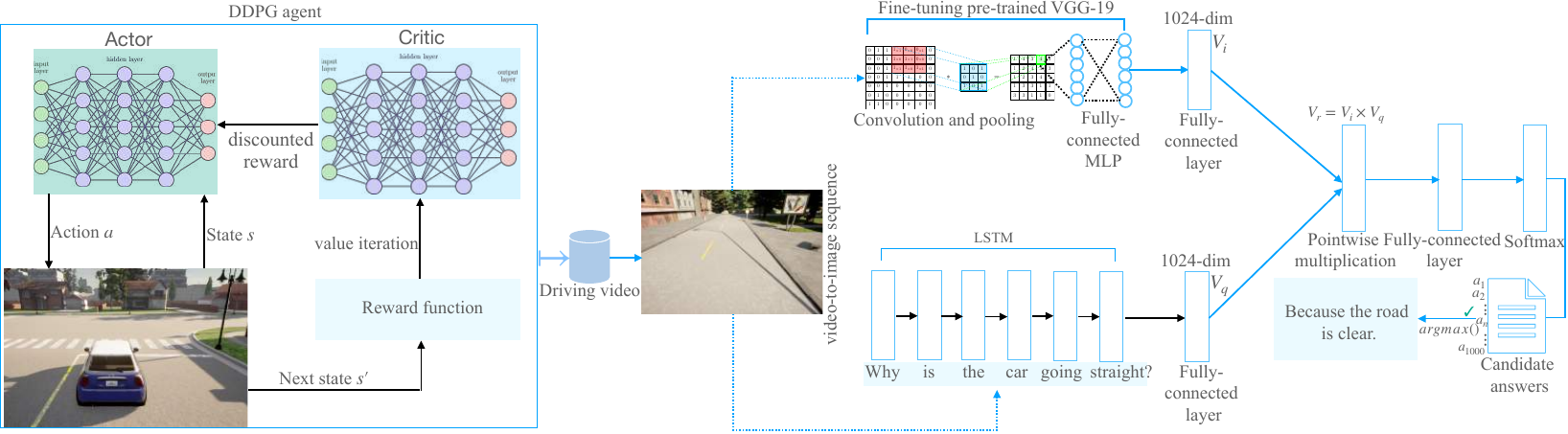}
    \caption{A diagram of the proposed VQA architecture for autonomous driving.}
    \label{fig:VQA_AD}
\end{figure*} \hspace{-0.50cm}
default, the waypoints are referenced to the origin point (0,0,0) in the map. To ensure that they are referenced to the dynamic position of the self-driving car while in motion, we use Perez et al.s’ methodology \cite{perez2022deep} and apply a transformation matrix
to represent the state of the agent  with these points, the 
vehicle’s yaw angle, and its global position on the map as follows:

\begin{equation}
 \begin{bmatrix}
\centering
\cos \phi_c & -\sin \phi_c & 0 & X_c\\
\sin \phi_c & \cos \phi_c & 0 & Y_c \\
0 & 0 & 1 & Z_c\\
0 & 0 & 1 & 1 
\end{bmatrix}   
\end{equation}
The goal of the task is that the ego car follows this predefined route and reaches the final destination by performing the relevant actions along its trip.\\
As seen from Figure \ref{fig:state_space}, the agent acquires a driving vector $f_t$= ($v_t$, $d_t$, $\phi_t$) from the simulation environment where these parameters reflect the vehicle's velocity, lateral distance, and yaw angle, respectively. Ideally, the goal of driving is to move on in the direction of the lane as long as possible without lane departure and collisions. In this sense, the reward shaping can be conditioned for the vehicle's 1) perfect longitudinal direction, 2) deviation from the lane direction with yaw angle, and 3) lane departure and collision. Based on these criteria, we adopt the relevant reward formulation from Perez et al. \cite{perez2022deep} for an ego car:

\begin{equation}
\resizebox{1\columnwidth}{!}{$R =
\begin{cases}
-200 & \text{road departures or collisions}, \\
\sum_{t}\left|v_{t} \cos \phi_{t}\right|-\left|v_{t} \sin \phi_{t}\right|-\left|v_{t}\right|\left|d_{t}\right| & \text{ driving inside the lane}, \\
+100 & \text{ arriving at the goal position}.
\end{cases}$}
\end{equation}

Finally, action space is continuous and can receive values from the interval  [-1,1]. By defining this setting, we trained our agent in Town 1. The training parameters of DDPG can be seen in Table \ref{Tab:tr_params}.
\begin{figure*}[htp!]
 \centering
    \includegraphics[width=17 cm]{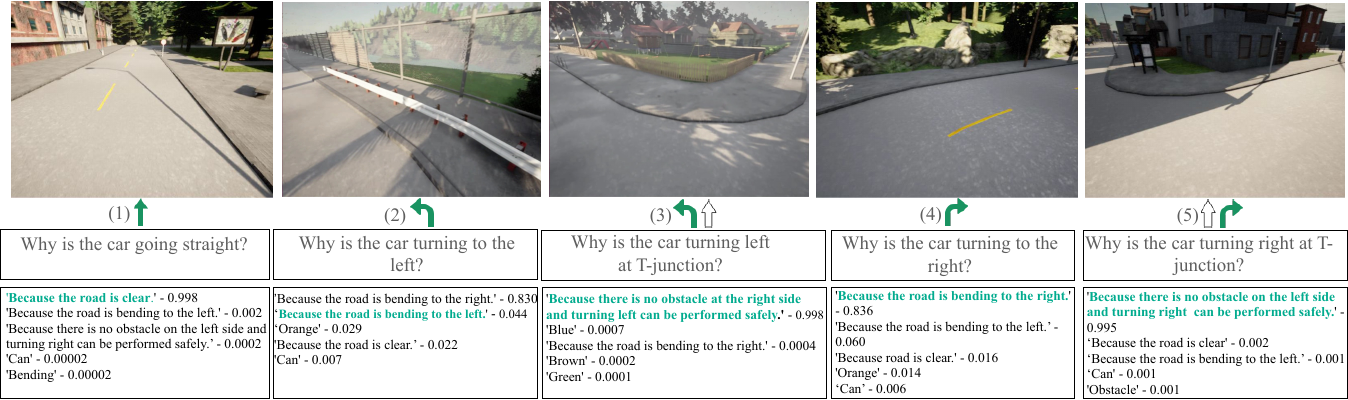}
    \caption{Example scenarios from an ego vehicle's field of view on CARLA. During the decision-making process of the agent, we are given visual signals and we ask action-related questions and try to find an answer given the current state. The green arrow shows the ego car's chosen action and the white arrows indicate the other route at T-junction scenarios. We show the top 5 answers predicted by our model. The green-colored text shows the correct answer to the question for the performed action of the car. Except for the \textit{turn left} scenario, justifications for other actions are predicted correctly by the model.}
    \label{fig:VQA_five_actions}
\end{figure*} \hspace{-0.46cm}

\begin{table*}[htbp]
  \centering
  \caption{Number of correct predictions for each action category}
  \label{tab:my_table}
  \begin{tabular*}{1\textwidth}{@{\extracolsep{\fill}}cccccc}
    \toprule
    \textbf{Go straight} & \textbf{Turn left} & \textbf{Turn left at T-junction} & \textbf{Turn right} & \textbf{Turn right at T-junction } & \textbf{Total }\\
    \midrule
    20/20 & 0/20 & 20/20 & 20/20 & 20/20 & 80/100 \\ 
    \bottomrule
    \label{Tab:eval}
  \end{tabular*}
\end{table*}
\subsubsection{Data Annotation}
As we obtained the driving video with the DDPG agent, we selected 5 action categories (\textit{go straight, turn left, turn right, turn left at T-junction, and turn right at T-junction}), and extracted consecutive frames uniformly (30 frames per second) for 5 video segments. We then chose 10 frames from each segment. We ensured that these frames were extracted from driving segments, where the car followed the predefined route and performed the relevant action safely without lane departure or collision. We distinguish left and right turns in the current line from left and right turns at T-junction as in the latter an ego car also has an alternative route. So, our training data includes 5 action categories with 50 high-quality frames per category, denoting a total of 250 driving scenes obtained from the recorded video. We manually annotated the training data with five causal question-answer (QA) pairs (see Table \ref{Tab:Tcr}) ensuring the annotations reflected the scene correctly. Each of 250 frames has its single and scenario-related annotation. As test data, we selected a collection of 100 frames from both Town 1 and Town 2 on the CARLA simulator, as the map of Town 2 is similar to Town 1. Similar to the training data annotation, we selected 20 frames for each action category from various segments of Town 1 and Town 2 and annotated each of them with a relevant QA pair. The goal is to assess the generalization ability of the employed VQA framework on these action categories in unseen traffic scenarios.
\subsubsection{Question-Answering Framework} On the question-answering side, we fine-tune the original VQA framework \cite{antol2015vqa} trained on the MS COCO dataset \cite{lin2014microsoft}. At the highest level, our VQA model takes an encoded driving image and a question embedding as input, to predict the answer (i.e., explanation) for a performed action in the scene. The model is composed of two neural networks. The first one is a multilayer-feedforward network with 2 hidden layers each containing 1000 hidden units and \textit{tanh} activation function. We apply the dropout regularization with 0.5 in each layer. Finally, a long short-term memory (LSTM) \cite{hochreiter1997long} followed by a softmax layer is employed to produce an answer for the asked question about the driving action. On the image encoding part, we eliminate the output layer and use the last hidden layer of the pre-trained VGG-19 architecture \cite{simonyan2015vgg}, producing a 4096-dimensional feature vector. Further, a linear transformation is applied to make the image features 1024-dimensional. The LSTM model for the question encoder has 2 hidden layers with 512 hidden units, and thus it is a 1024-dimensional vector, the same as image features. An interesting aspect is  the unification of the question and image vectors from a mathematical perspective. Previous studies have generally either preferred the concatenation or multiplication of these vectors, but \cite{antol2015vqa} and \cite{ garg2018object} have shown that multiplying the image and question encoder usually leads to a better joint representation. Consequently, given the image vector, $V_i$, and question embedding $V_q$, the resulting vector passed to the fully connected layer of the VQA pipeline is represented as their element-wise multiplication, as a fused feature vector:

\begin{equation}
V_r= V_i \times V_q
\end{equation}

We use the question and answer vocabularies of the original VQA framework, which have sizes of more than 17K unique tokens and 1000 candidate answers (which are either single tokens such as ``yes,'' ``white,'' or expressions consisting of two or more strings such as ``playing video game''), respectively, obtained by descriptions from the MS COCO \cite{lin2014microsoft} images. We customize candidate answers by adding our answers of 5 action questions to that answer vocabulary. The expectation is that our VQA model picks the most correct answer with the highest softmax probability score out of the 1K candidates for the asked ``Why'' question about the action within the driving scene.

\section{Experimental Results and Discussion}
On the data collection side, we trained the DDPG agent on the CARLA 0.9.11 version in 500 episodes using a TensorFlow backend to get a driving video. As described above, we used 250 frames from Town 1 for training our VQA network and evaluated its performance on 100 frames collected from Town 1 and Town 2 (Figure \ref{fig:towns}). We used the PyTorch backend for training and evaluating our VQA architecture. The experiments were performed on an NVIDIA RTX 3090 GPU machine with a 32 GB memory size. All the frames were set to have a size of 640 $\times$ 480 both in training and test. As we have ground-truth answers (see Table \ref{Tab:Tcr}) for the asked question about an image, we compare the top prediction of our model on the test data (i.e., an answer with the highest softmax probability score) with these ground-truth answers. Thus, we use accuracy as an evaluation metric, which is defined as follows:

\begin{figure}[t!]
\centering
\includegraphics[width = 0.9\columnwidth]{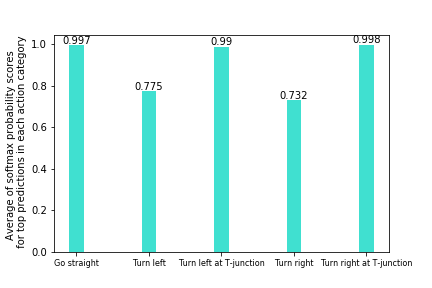}
\caption{The average softmax probability scores for top predictions in each action category.}
\label{fig:softmax_scores}
\end{figure}
\begin{equation}
    \text{Accuracy} = \frac{ \# ~ frames ~ with ~ correct ~ predictions}{total ~ number ~ of ~ test ~ frames  }
\end{equation}

Based on this evaluation criterion, our VQA model predicted 80 correct answers to the asked questions for 100 images. Hence, the accuracy of the prediction is 0.8 or 80\%.
\subsubsection{Discussion} Except for \textit{turn left} actions, our model predicted explanatory answers correctly for all remaining action classes. Interestingly, in frames with turn-left scenarios, the VQA framework primarily recognized these actions as \textit{turn right}. In Figure \ref{fig:VQA_five_actions}, we provide exemplary driving scenes for the five action categories. As seen, the model was able to predict the highest probability scores for all actions in the scenes correctly, except for the misclassified \textit{turn left} action in the second image. This misclassification could be due to ambiguity in the tested driving frames, the shape of curves in the scene, and road conditions in the training data. Hence, it is important to increase the size of the training data considering the shapes of road lanes and curves, lighting, and other road objects to potentially improve the accuracy of the predictions of the VQA network on self-driving actions.\\
Another implication of our work is that unifying computer vision with a natural language provides an opportunity to explain temporal actions of an RL agent.   As explored in a  recent study \cite{peng2022inherently}, explaining RL in sequential decision-making problems is an important and emerging topic, particularly when explanation receivers do not have a technical background. As autonomous driving is a safety-critical application area, justifying reinforcement learning-based decisions to end users with natural language-based reasoning is an effective and easily understandable approach. A natural foundation for explainable reinforcement learning (XRL) would be to provide reward-based justifications on action decisions. However, as self-driving explanations are intended for a general community, it is essential to ensure that such explanations are intelligible and informative. While \cite{peng2022inherently} has attempted to build an inherently explainable RL architecture, we build our explanations independent of an agent's decisions. We also acknowledge the need to be cautious about providing explanations that are independent of an agent's behavior; it is possible that post-hoc explanations may {\it not} always reflect an agent's real decision-making process. For example, in an actual \textit{left turn} scenario, a model's response to the question ``Why is the car turning to the \textit{right}?" as "Because the road is bending to the \textit{right}." may be a hallucination of a VQA architecture. Consequently, it is important to further investigate the topic of generating linguistic explanations for an agent's actions and evaluate such explanations with human-adversarial examples as well.  
\subsubsection{Limitations}
Real roads are more complex and dynamic with the presence of traffic lights, bystanders, passengers, other vehicles, and adverse weather conditions. In the current version of our framework, the ego car only interacts with the stationary environment and explains actions associated with such interactions. Moreover, our dataset is small in size. Hence, these features are limitations of our present framework, and as a next step, we plan to work on explaining self-driving actions in more dynamic and complex scenarios with enriched data, where details are provided in the conclusions section.
\subsubsection{Practical use cases} In practice, the VQA mechanism can be leveraged at least in two ways on real autonomous vehicles. First, it can help passengers on board monitor driving safety by ``judging'' the vehicle's decisions. For instance, a user interface or dashboard set up on a back seat may provide voice-to-text functionality, and a passenger can observe driving surrounding, ask a question about the vehicle's chosen action, and get an answer. Such a feature can help monitor the reliability of self-driving and instill trust in vehicle autonomy during the trip. Another practical application is to retain a history of action-question-answer triplets (...$a_t, q_t, ans_t, a_{t+1}, q_{t+1}, ans_{t+1}...$) and use it for forensic analysis in possible accident investigations with self-driving vehicles. Such log data can help understand why the self-driving vehicle made a specific decision at a particular time just before being involved in an accident. 
\section{Conclusions and Future Work}
We have presented a preliminary study on explaining autonomous driving actions with a VQA approach. We used driving data generated by an RL agent on the CARLA simulator and developed our question-answering system as an explanatory approach to the agent's decisions. The experimental results show that a simple and straightforward VQA mechanism can help interpret the real-time decisions of an autonomous car and also help understand its correct and incorrect decisions as safety implications. The results also suggest that unifying VQA with RL-based decision-making will likely do well for actions in a dynamic environment, provided that we have more training dataset. In this sense, we plan to explore three potential directions: \\
 \textit{1. Augmenting data and using other VQA architectures:} We will increase the size of training data (ideally $>$50K driving frames), perform fine-tuning on our model using more recent ConvNet architectures, such as Vision Transformer (ViT) \cite{dosovitskiy2021an}, and try out other VQA frameworks as well. By making a comparative analysis of these pre-trained deep neural architectures on driving data, we can observe their empirical performance in terms of accuracy and potentially produce a large-scale and curated benchmark dataset.  \\
 \textit{2. Training an RL agent on dynamic environments:} We will run the RL agent in other towns on the CARLA simulator, which have more vehicles, pedestrians, and complex intersections, and annotate the ego vehicle's interaction with them accordingly to provide more image-question-answer triplets.  \\
 \textit{3. Leveraging large language models (LLMs):} Finally, a recent breakthrough in LLMs gives a reason to use this architecture in autonomous driving problems. As our current task combines vision and natural language-based reasoning for explaining self-driving actions, multimodal transformers (e.g., GPT-4 \cite{openai2023gpt4} and its variations) could serve a purpose in this context. As multimodal transformers can input an image and text, and provide contextual information about the joint semantics, it seems promising to fine-tune such state-of-the-art learning architectures for the self-driving domain and generate rigorously structured explanations. \\ 
 We believe that the empirical work and further directions proposed in this paper can help improve safety, transparency, and trustworthiness of autonomous driving technology.
\bibliography{IEEEabrv.bib, main.bib}{}

\begin{thebibliography}{10}
\providecommand{\url}[1]{#1}
\csname url@rmstyle\endcsname
\providecommand{\newblock}{\relax}
\providecommand{\bibinfo}[2]{#2}
\providecommand\BIBentrySTDinterwordspacing{\spaceskip=0pt\relax}
\providecommand\BIBentryALTinterwordstretchfactor{4}
\providecommand\BIBentryALTinterwordspacing{\spaceskip=\fontdimen2\font plus
\BIBentryALTinterwordstretchfactor\fontdimen3\font minus
  \fontdimen4\font\relax}
\providecommand\BIBforeignlanguage[2]{{%
\expandafter\ifx\csname l@#1\endcsname\relax
\typeout{** WARNING: IEEEtran.bst: No hyphenation pattern has been}%
\typeout{** loaded for the language `#1'. Using the pattern for}%
\typeout{** the default language instead.}%
\else
\language=\csname l@#1\endcsname
\fi
#2}}

\bibitem{waymo_one_million}
\BIBentryALTinterwordspacing
{Waymo's Blog}, ``{First Million Rider-Only Miles: How the Waymo Driver is
  Improving Road Safety},'' { Accessed online} on March 12, 2023. [Online].
  Available:
  \url{https://blog.waymo.com/2023/02/first-million-rider-only-miles-how.html}
\BIBentrySTDinterwordspacing

\bibitem{board2020collision}
{NTS Board}, ``Collision between a sport utility vehicle operating with partial
  driving automation and a crash attenuator {Mountain View}, {California.}'' {
  Accessed online} on March 9, 2023.

\bibitem{voigt2017eu}
P.~Voigt and A.~Von~dem Bussche, ``{The EU General Data Protection Regulation
  (GDPR)},'' \emph{A Practical Guide, 1st Ed., Cham: Springer International
  Publishing}, vol.~10, no. 3152676, pp. 10--5555, 2017.

\bibitem{gdpr_article_22}
\BIBentryALTinterwordspacing
{GDPR}, ``{Art. 22 GDPR Automated individual decision-making, including
  profiling },'' (Accessed on March 9, 2023). [Online]. Available:
  \url{https://gdpr-info.eu/art-22-gdpr/}
\BIBentrySTDinterwordspacing

\bibitem{omeiza2021explanations}
D.~Omeiza, H.~Webb, M.~Jirotka, and L.~Kunze, ``{Explanations in Autonomous
  Driving: A Survey},'' \emph{IEEE Transactions on Intelligent Transportation
  Systems}, vol.~23, no.~8, pp. 10\,142--10\,162, 2021.

\bibitem{atakishiyev2021explainable}
S.~Atakishiyev, M.~Salameh, H.~Yao, and R.~Goebel, ``{Explainable Artificial
  Intelligence for Autonomous Driving: A Comprehensive Overview and Field Guide
  for Future Research Directions},'' \emph{arXiv preprint arXiv:2112.11561},
  2021.

\bibitem{zablocki2022explainability}
{\'E}.~Zablocki, H.~Ben-Younes, P.~P{\'e}rez, and M.~Cord, ``{Explainability of
  Deep Vision-Based Autonomous Driving Systems: Review and Challenges},''
  \emph{International Journal of Computer Vision}, vol. 130, no.~10, pp.
  2425--2452, 2022.

\bibitem{kafle2017visual}
K.~Kafle and C.~Kanan, ``Visual question answering: Datasets, algorithms, and
  future challenges,'' \emph{Computer Vision and Image Understanding}, vol.
  163, pp. 3--20, 2017.

\bibitem{bojarski2016end}
M.~Bojarski, D.~Del~Testa, D.~Dworakowski, B.~Firner, B.~Flepp, P.~Goyal, L.~D.
  Jackel, M.~Monfort, U.~Muller, J.~Zhang, \emph{et~al.}, ``End to end learning
  for self-driving cars,'' \emph{arXiv preprint arXiv:1604.07316}, 2016.

\bibitem{kim2017interpretable}
J.~Kim and J.~Canny, ``{Interpretable Learning for Self-Driving Cars by
  Visualizing Causal Attention},'' in \emph{Proceedings of the IEEE
  International Conference on Computer Vision}, 2017, pp. 2942--2950.

\bibitem{bojarski2018visualbackprop}
M.~Bojarski, A.~Choromanska, K.~Choromanski, B.~Firner, L.~J. Ackel, U.~Muller,
  P.~Yeres, and K.~Zieba, ``{VisualBackProp: Efficient Visualization of CNNs
  for Autonomous Driving},'' in \emph{{2018 IEEE International Conference on
  Robotics and Automation (ICRA)}}.\hskip 1em plus 0.5em minus 0.4em\relax
  IEEE, 2018, pp. 4701--4708.

\bibitem{mohseni2019predicting}
S.~Mohseni, A.~Jagadeesh, and Z.~Wang, ``Predicting model failure using
  saliency maps in autonomous driving systems,'' \emph{ICML 2019 Workshop on
  Uncertainty and Robustness in Deep Learning}, 2019.

\bibitem{bansal2019chauffeurnet}
M.~Bansal, A.~Krizhevsky, and A.~Ogale, ``{ChauffeurNet: Learning to Drive by
  Imitating the Best and Synthesizing the Worst},'' \emph{Robotics: Science and
  Systems}, 2019.

\bibitem{li2020make}
C.~Li, S.~H. Chan, and Y.-T. Chen, ``{Who Make Drivers Stop? Towards
  Driver-centric Risk Assessment: Risk Object Identification via Causal
  Inference},'' in \emph{2020 IEEE/RSJ International Conference on Intelligent
  Robots and Systems (IROS)}.\hskip 1em plus 0.5em minus 0.4em\relax IEEE,
  2020, pp. 10\,711--10\,718.

\bibitem{dosovitskiy2017carla}
A.~Dosovitskiy, G.~Ros, F.~Codevilla, A.~Lopez, and V.~Koltun, ``{CARLA: An
  Open Urban Driving Simulator},'' in \emph{Conference on Robot
  Learning}.\hskip 1em plus 0.5em minus 0.4em\relax PMLR, 2017, pp. 1--16.

\bibitem{jacob2022steex}
P.~Jacob, {\'E}.~Zablocki, H.~Ben-Younes, M.~Chen, P.~P{\'e}rez, and M.~Cord,
  ``{STEEX: Steering Counterfactual Explanations with Semantics},'' in
  \emph{Proceedings of the European Conference on Computer Vision
  (ECCV)}.\hskip 1em plus 0.5em minus 0.4em\relax Springer, 2022, pp. 387--403.

\bibitem{kim2018textual}
J.~Kim, A.~Rohrbach, T.~Darrell, J.~Canny, and Z.~Akata, ``{Textual
  Explanations for Self-Driving Vehicles},'' in \emph{Proceedings of the
  European Conference on Computer Vision (ECCV)}, 2018, pp. 563--578.

\bibitem{kim2019grounding}
J.~Kim, T.~Misu, Y.-T. Chen, A.~Tawari, and J.~Canny, ``{Grounding
  Human-to-Vehicle Advice for Self-driving Vehicles},'' in \emph{{Proceedings
  of the IEEE/CVF Conference on Computer Vision and Pattern Recognition}},
  2019, pp. 10\,591--10\,599.

\bibitem{kim2020advisable}
J.~Kim, S.~Moon, A.~Rohrbach, T.~Darrell, and J.~Canny, ``{Advisable Learning
  for Self-Driving Vehicles by Internalizing Observation-to-Action Rules},'' in
  \emph{Proceedings of the IEEE/CVF Conference on Computer Vision and Pattern
  Recognition}, 2020, pp. 9661--9670.

\bibitem{xu2020explainable}
Y.~Xu, X.~Yang, L.~Gong, H.-C. Lin, T.-Y. Wu, Y.~Li, and N.~Vasconcelos,
  ``{Explainable Object-induced Action Decision for Autonomous Vehicles},'' in
  \emph{Proceedings of the IEEE/CVF Conference on Computer Vision and Pattern
  Recognition}, 2020, pp. 9523--9532.

\bibitem{ben2022driving}
H.~Ben-Younes, {\'E}.~Zablocki, P.~P{\'e}rez, and M.~Cord, ``Driving behavior
  explanation with multi-level fusion,'' \emph{Pattern Recognition}, vol. 123,
  p. 108421, 2022.

\bibitem{brewitt2021grit}
C.~Brewitt, B.~Gyevnar, S.~Garcin, and S.~V. Albrecht, ``{GRIT: Fast,
  Interpretable, and Verifiable Goal Recognition with Learned Decision Trees
  for Autonomous Driving},'' in \emph{2021 IEEE/RSJ International Conference on
  Intelligent Robots and Systems (IROS)}.\hskip 1em plus 0.5em minus
  0.4em\relax IEEE, 2021, pp. 1023--1030.

\bibitem{almalioglu2022deep}
Y.~Almalioglu, M.~Turan, N.~Trigoni, and A.~Markham, ``Deep learning-based
  robust positioning for all-weather autonomous driving,'' \emph{Nature Machine
  Intelligence}, vol.~4, no.~9, pp. 749--760, 2022.

\bibitem{lee2019attitudes}
D.~Lee, J.~Mulrow, C.~J. Haboucha, S.~Derrible, and Y.~Shiftan, ``{Attitudes on
  Autonomous Vehicle Adoption using Interpretable Gradient Boosting Machine},''
  \emph{Transportation Research Record}, vol. 2673, no.~11, pp. 865--878, 2019.

\bibitem{schneider2021increasing}
T.~Schneider, S.~Ghellal, S.~Love, and A.~R. Gerlicher, ``{Increasing the User
  Experience in Autonomous Driving through different Feedback Modalities},'' in
  \emph{26th International Conference on Intelligent User Interfaces}, 2021,
  pp. 7--10.

\bibitem{schneider2021explain}
T.~Schneider, J.~Hois, A.~Rosenstein, S.~Ghellal, D.~Theofanou-F{\"u}lbier, and
  A.~R. Gerlicher, ``{ExplAIn Yourself! Transparency for Positive UX in
  Autonomous Driving},'' in \emph{Proceedings of the 2021 CHI Conference on
  Human Factors in Computing Systems}, 2021, pp. 1--12.

\bibitem{gurari2018vizwiz}
D.~Gurari, Q.~Li, A.~J. Stangl, A.~Guo, C.~Lin, K.~Grauman, J.~Luo, and J.~P.
  Bigham, ``{VizWiz Grand Challenge: Answering Visual Questions From Blind
  People},'' in \emph{Proceedings of the IEEE Conference on Computer Vision and
  Pattern Recognition}, 2018, pp. 3608--3617.

\bibitem{zhan2020medical}
L.-M. Zhan, B.~Liu, L.~Fan, J.~Chen, and X.-M. Wu, ``{Medical Visual Question
  Answering via Conditional Reasoning},'' in \emph{Proceedings of the 28th ACM
  International Conference on Multimedia}, 2020, pp. 2345--2354.

\bibitem{li2019isee}
D.~Li, Z.~Zhang, K.~Yu, K.~Huang, and T.~Tan, ``{ISEE: An Intelligent Scene
  Exploration and Evaluation Platform for Large-Scale Visual Surveillance},''
  \emph{IEEE Transactions on Parallel and Distributed Systems}, vol.~30,
  no.~12, pp. 2743--2758, 2019.

\bibitem{Yi2020CLEVRER}
K.~Yi, C.~Gan, Y.~Li, P.~Kohli, J.~Wu, A.~Torralba, and J.~B. Tenenbaum,
  ``{CLEVRER: Collision Events for Video Representation and Reasoning},'' in
  \emph{International Conference on Learning Representations}, 2020.

\bibitem{xu2021sutd}
L.~Xu, H.~Huang, and J.~Liu, ``{SUTD-TrafficQA: A Question Answering Benchmark
  and an Efficient Network for Video Reasoning Over Traffic Events},'' in
  \emph{Proceedings of the IEEE/CVF Conference on Computer Vision and Pattern
  Recognition}, 2021, pp. 9878--9888.

\bibitem{lillicrap2016}
T.~P. Lillicrap, J.~J. Hunt, A.~Pritzel, N.~Heess, T.~Erez, Y.~Tassa,
  D.~Silver, and D.~Wierstra, ``{Continuous Control with Deep Reinforcement
  Learning},'' \emph{International Conference on Learning Representations},
  2016.

\bibitem{hart1968formal}
P.~E. Hart, N.~J. Nilsson, and B.~Raphael, ``{A Formal Basis for the Heuristic
  Determination of Minimum Cost Paths},'' \emph{IEEE Transactions on Systems
  Science and Cybernetics}, vol.~4, no.~2, pp. 100--107, 1968.

\bibitem{perez2022deep}
{\'O}.~P{\'e}rez-Gil, R.~Barea, E.~L{\'o}pez-Guill{\'e}n, L.~M. Bergasa,
  C.~Gomez-Huelamo, R.~Guti{\'e}rrez, and A.~Diaz-Diaz, ``{Deep reinforcement
  learning based control for Autonomous Vehicles in CARLA},'' \emph{Multimedia
  Tools and Applications}, vol.~81, no.~3, pp. 3553--3576, 2022.

\bibitem{antol2015vqa}
S.~Antol, A.~Agrawal, J.~Lu, M.~Mitchell, D.~Batra, C.~L. Zitnick, and
  D.~Parikh, ``{VQA: Visual Question Answering},'' in \emph{Proceedings of the
  IEEE International Conference on Computer Vision}, 2015, pp. 2425--2433.

\bibitem{lin2014microsoft}
T.-Y. Lin, M.~Maire, S.~Belongie, J.~Hays, P.~Perona, D.~Ramanan,
  P.~Doll{\'a}r, and C.~L. Zitnick, ``{Microsoft COCO: Common Objects in
  Context},'' in \emph{Computer Vision--ECCV 2014: 13th European Conference,
  Zurich, Switzerland, September 6-12, 2014, Proceedings, Part V 13}.\hskip 1em
  plus 0.5em minus 0.4em\relax Springer, 2014, pp. 740--755.

\bibitem{hochreiter1997long}
S.~Hochreiter and J.~Schmidhuber, ``Long short-term memory,'' \emph{Neural
  computation}, vol.~9, no.~8, pp. 1735--1780, 1997.

\bibitem{simonyan2015vgg}
K.~Simonyan and A.~Zisserman, ``{Very Deep Convolutional Networks for
  Large-Scale Image Recognition},'' \emph{International Conference on Learning
  Representations}, 2015.

\bibitem{garg2018object}
S.~Garg and R.~Srivastava, ``Object sequences: encoding categorical and spatial
  information for a yes/no visual question answering task,'' \emph{IET Computer
  Vision}, vol.~12, no.~8, pp. 1141--1150, 2018.

\bibitem{peng2022inherently}
X.~Peng, M.~Riedl, and P.~Ammanabrolu, ``{Inherently Explainable Reinforcement
  Learning in Natural Language},'' \emph{Advances in Neural Information
  Processing Systems}, vol.~35, pp. 16\,178--16\,190, 2022.

\bibitem{dosovitskiy2021an}
A.~Dosovitskiy, L.~Beyer, A.~Kolesnikov, D.~Weissenborn, X.~Zhai,
  T.~Unterthiner, M.~Dehghani, M.~Minderer, G.~Heigold, S.~Gelly, J.~Uszkoreit,
  and N.~Houlsby, ``{An Image is Worth 16x16 Words: Transformers for Image
  Recognition at Scale},'' in \emph{International Conference on Learning
  Representations}, 2021.

\bibitem{openai2023gpt4}
{OpenAI}, ``{GPT-4 Technical Report},'' \url{https://arxiv.org/abs/2303.08774},
  2023.

\end{thebibliography}
\bibliographystyle{IEEEtran}

\end{document}